\documentclass[preprint,12pt]{elsarticle}

\usepackage{epsfig}

\usepackage{amssymb}

\usepackage{amsmath}
\usepackage{amsthm}
\usepackage{caption}
\usepackage{subcaption}
\usepackage{mathabx}

\newcommand{\ignore}[1]{}

\usepackage{lipsum}
\makeatletter
\def\ps@pprintTitle{%
 \let\@oddhead\@empty
 \let\@evenhead\@empty
 \def\@oddfoot{}%
 \let \@oddfoot}
\makeatother
\begin{document}

\begin{frontmatter}



\title{A Two-Stage Shape Retrieval (TSR) Method with Global and Local Features}


\author{Xiaqing Pan}
\author{Sachin Chachada}
\author{C.-C. Jay Kuo}
\address{Department of Electrical Engineering, University of Southern California, Los Angeles, CA 90089-2564, USA}

\begin{abstract}
A robust two-stage shape retrieval (TSR) method is proposed to address
the 2D shape retrieval problem. Most state-of-the-art shape retrieval
methods are based on local features matching and ranking. Their
retrieval performance is not robust since they may retrieve globally
dissimilar shapes in high ranks. To overcome this challenge, we
decompose the decision process into two stages. In the first irrelevant
cluster filtering (ICF) stage, we consider both global and local features
and use them to predict the relevance of gallery shapes with respect to
the query. Irrelevant shapes are removed from the candidate shape set.
After that, a local-features-based matching and ranking (LMR) method
follows in the second stage.  We apply the proposed TSR system to
MPEG-7, Kimia99 and Tari1000 three datasets and show that it outperforms
all other existing methods.  The robust retrieval performance of the TSR
system is demonstrated.
\end{abstract}


\end{frontmatter}


\section{Introduction}\label{sec.introduction}

2D shapes, also known as silhouette images, are often encountered in
computer vision tasks such as manufacture components recognition and
retrieval, sketch-based shape retrieval, medical image analysis, etc.
Given a 2D shape as the query, a shape retrieval system retrieves ranked
shapes from a gallery set according to a certain similarity measure
between the query shape and shapes in the retrieval dataset, called
gallery shapes. The performance is evaluated by consistency between
ranked shapes and human interpretation.  The 2D shape retrieval problem
is challenging due to a wide range of shape variations, including
articulation, noise, contour deformation, topological transformation and
multiple projection angles. It is worthwhile to emphasize that our
research addresses a retrieval problem with no labeled data at all.  It
is very different from the deep learning architecture, such as that in
\cite{krizhevsky2012imagenet}, that relies on a huge amount of labeled
data for training. 

Traditionally, the similarity between two shapes is measured using
global or local features that capture shape properties such as contours
and regions. Global features include the Zernike moment
\cite{khotanzad1990invariant} and the Fourier descriptor
\cite{zhang2002shape}. They are however not effective in capturing local
details of shape contours, resulting in low discriminative performance.
Recent research efforts have focused on the development of more powerful
local features and post-processing techniques ({\em e.g.}, diffusion). A
substantial progress has been made in this area and will be briefly
reviewed below.

\begin{figure}[!th]
\centering
\begin{subfigure}[b] {\linewidth}
\centering
\includegraphics[width=0.9\linewidth]{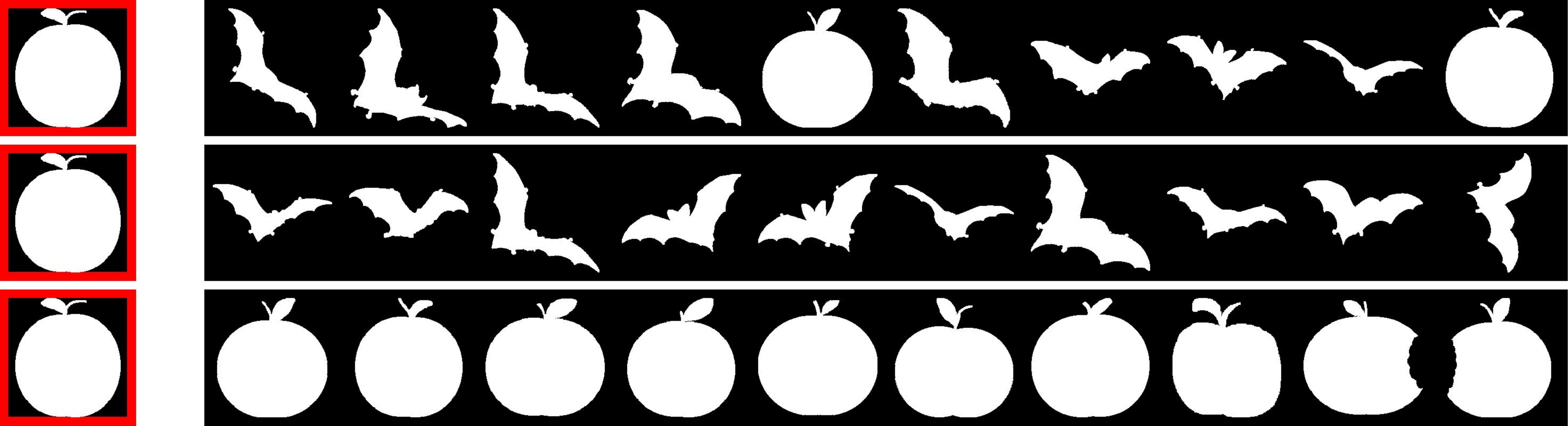}
\caption{Apple}
\end{subfigure}
\centering
\begin{subfigure}[b]{\linewidth}
\centering
\includegraphics[width=0.9\linewidth]{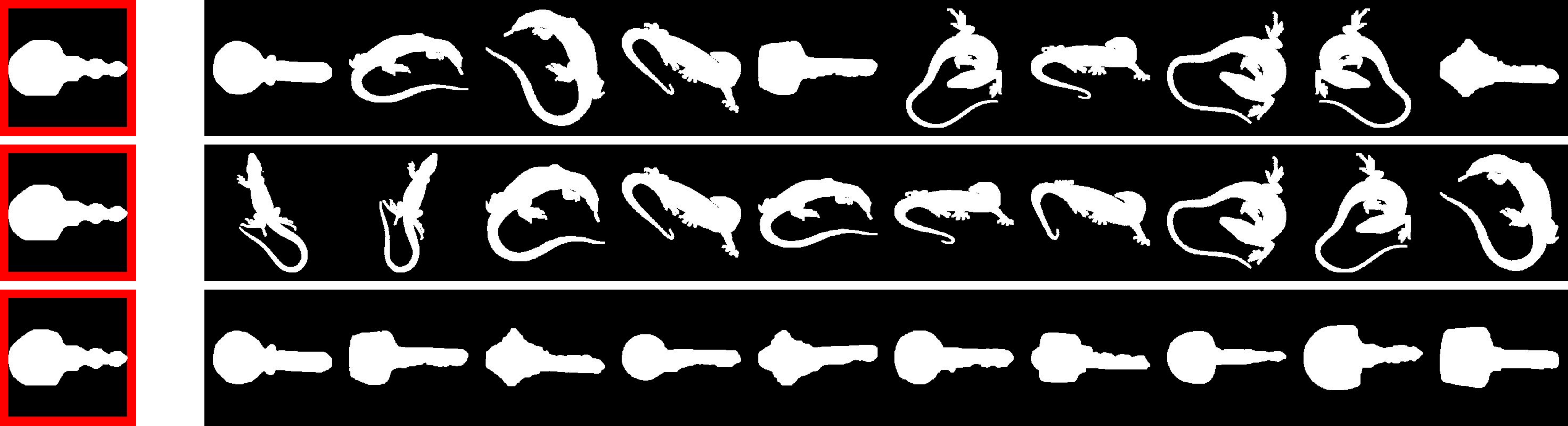}
\caption{Key}
\end{subfigure}
\centering
\begin{subfigure}[b]{\linewidth}
\centering
\includegraphics[width=0.9\linewidth]{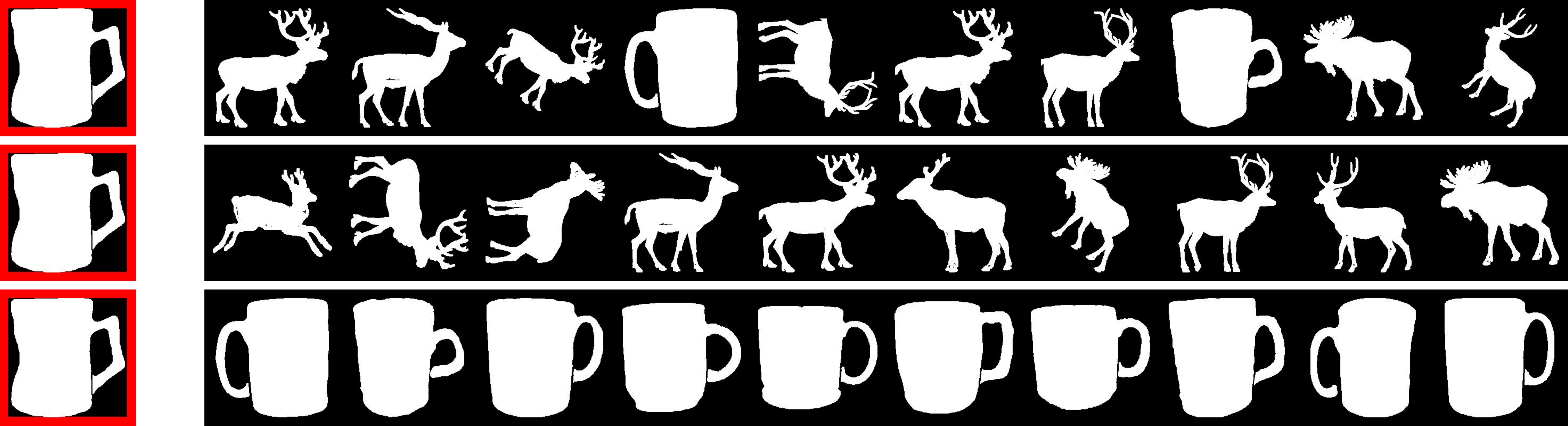}
\caption{Cup}
\end{subfigure}
\centering
\begin{subfigure}[b]{\linewidth}
\centering
\includegraphics[width=0.9\linewidth]{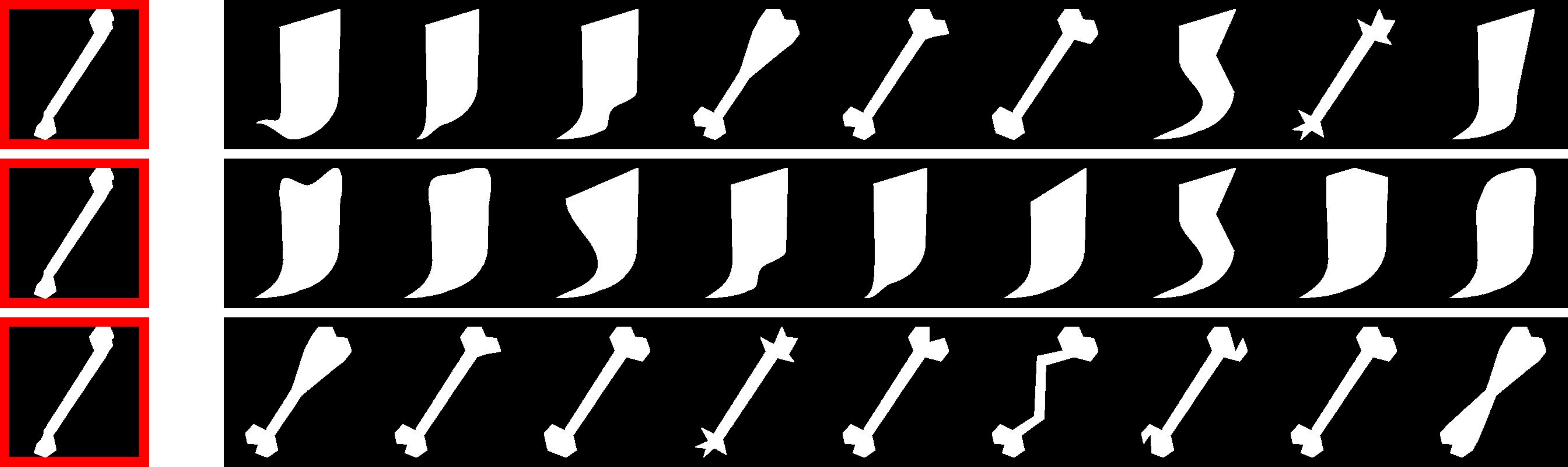}
\caption{Bone}
\end{subfigure}
\caption{Comparison of retrieved shapes using AIR (the first row),
AIR+DP (the second row) and the proposed TSR method (the third row)
with respect to four query shapes: (a) Apple, (b) Key, (c) Cup,
and (d) Bone.}\label{fig.irrelevant}
\end{figure}

The shape context (SC) method \cite{belongie2002shape} describes a
contour point by its relationship to other contour points in a local log
polar coordination.  However, the Euclidian distance used to construct a
local coordination is sensitive to articulation variations.  The inner-distance shape context (IDSC) method \cite{ling2007shape} attempts to
resolve the articulation problem by using the inner-distance between two
points on a contour. The aspect shape context (ASC) method
\cite{ling2010balancing} and the articulation invariant representation
(AIR) method \cite{gopalan2010articulation} extend IDSC to account for
shape interior variations and projection variations, respectively.  Fast
computation of the elastic geodesic distance in
\cite{srivastava2011shape} for shape retrieval is recently studied in
\cite{dogan2015fast}.  Although local-features-based methods capture
important shape properties, their locality restricts discrimination
among shapes on the global scale. Consequently, their retrieval results
may include globally irrelevant shapes in high ranks.  To illustrate
this claim, three exemplary query shapes, given in the leftmost column
of each row, and their top 10-ranked retrieval results are displayed
from left to right in rank order in Figs.  \ref{fig.irrelevant}(a)-(d).
The results of the AIR method are shown in the first row of each
subfigure.  Apparently, these retrieved results are against human
intuition.

Post-processing techniques such as the diffusion process (DP)
\cite{donoser2013diffusion}, \cite{premachandran2013consensus},
\cite{qadeer2015improving}, \cite{yang2009locally},
\cite{yang2013affinity} have been proposed to compensate for errors arising
from local-features-based shape retrieval methods.  The DP treats each
sample as a node and the similarity between any two samples corresponds
to a weighted edge. All samples form a connected graph, called a
manifold, and affinities are diffused along the manifold to improve
measured similarities.  However, the DP has its limitations.  When
shapes of two classes are mixed in the feature space, it cannot separate
them properly. Also, when a query is far away from the majority of samples in its
class, it is difficult to retrieve shapes of the same class with high
ranks. For example, AIR is confused with several classes as shown in the
first row of Figs.  \ref{fig.irrelevant}(a)-(d). Even the best DP does
not help retrieval results much as shown in the second row of Figs.
\ref{fig.irrelevant} (a)-(d).  Clearly, the DP is constrained by the
underlying feature space.

Being motivated by the above observations, we develop a more robust
shape retrieval system with two main contributions in this paper.
First, we consider both global and local features. Since traditional
global features are not discriminative enough, we develop more powerful
and robust global features. Second, we propose a two-stage shape
retrieval (TSR) system that consists of: I) the irrelevant cluster
filtering (ICF) stage and II) the local-features-based matching and
ranking (LMR) stage. For Stage II, we can adopt any state-of-the-art
shape retrieval solution, and our focus is on the design of Stage I.
The robustness of the proposed TSR system can be intuitively explained
below. In the ICF stage (Stage I), we attempt to cluster gallery shapes
that are similar to each other by examining global and local features
simultaneously.  Two contradictory cases may arise. Shapes that are
close in the local feature space can be distant in the global feature
space, and vice versa. Here, we resolve the contradiction with a joint
cost function that strikes a balance between the two distances.  It is
convenient to use the ICF stage to filter out unlikely shapes.  In
particular, shapes that are close in the local feature space but distant
in the global feature can be removed in this stage. Then, the TSR system
will avoid the same mistake of traditional one-stage matching methods
when it proceeds to its second LMR stage. The retrieved results of the
TSR system are shown in the third row of Figs. \ref{fig.irrelevant}
(a)-(d). All wrongly retrieved results are corrected by TSR.  The
novelty of our work lies in two areas: 1) identifying the cause of
unreliable retrieval results in all state-of-the-art methods for the 2D
shape retrieval problem, and 2) finding a new system-level framework to
solve this problem. 

The rest of this paper is organized as follows.  The TSR method is
described in Section \ref{sec2}.  Experimental results are shown in
Section \ref{sec.experiment}. Finally, concluding remarks are given in
Section \ref{sec.conclusion}. 

\section{Proposed TSR Method}\label{sec2}

\subsection{System Overview}

\begin{figure*}
\begin{center}
\includegraphics[width=\linewidth]{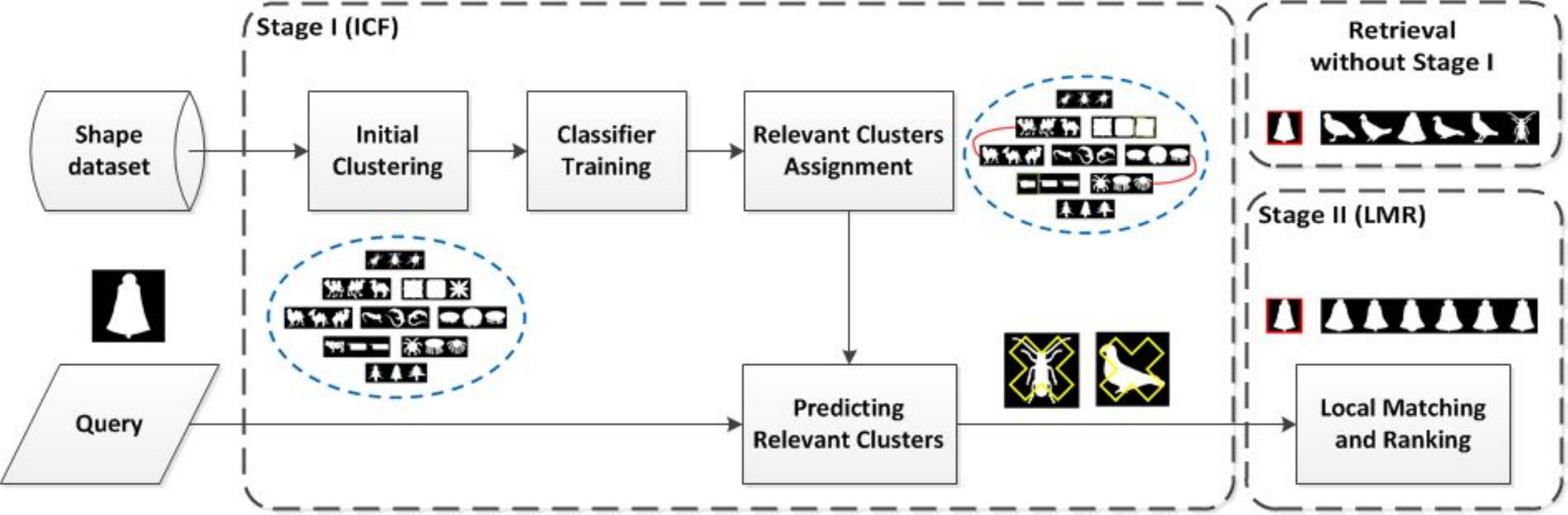}
\caption{The flow chart of the proposed TSR system.}\label{fig.flowchart}
\end{center}
\end{figure*}

An overview flow chart of the proposed TSR system is given in Fig.
\ref{fig.flowchart}. As shown in the figure, the system consists of two
stages. Stage I of the TSR system is trained in an off-line process with
the following three steps.
\begin{enumerate}
\itemsep -1ex
\item {\bf Initial clustering}. All samples in the dataset are clustered
using their local features.\\
\item {\bf Classifier Training.} Samples close to the centroid of each
cluster are selected as training data. Their extracted global features
are used to train a random forest classifier.\\
\item {\bf Relevant Clusters Assignment.} The trained random forest
classifier assigns relevant clusters to all samples in the dataset
so that each sample is associated with a small set of relevant clusters.
\end{enumerate}
In the on-line query process, we extract both global and local features
from a query shape and, then, proceed with the following two steps:
\begin{enumerate}
\itemsep -1ex
\item {\bf Predicting Relevant Clusters.} Given a query sample, we
assign it a set of relevant clusters based on a cost function. The cost function consists of two negative log likelihood terms. One likelihood
reflects the relevant cluster distribution of the query sample itself
while the other is the mean of the relevant cluster distributions of its
local neighbors.  The ultimate relevant clusters are obtained by
thresholding the cost function.\\
\item {\bf Local Matching and Ranking.} We conduct matching and ranking
for samples in the relevant clusters with a distance in the local
feature space. The diffusion process can also be applied to enhance
the retrieval accuracy.
\end{enumerate}

A query bell shape is given in Fig. \ref{fig.flowchart}. The traditional
shape retrieval algorithm (with Stage II only) finds birds, a bell and a
beetle in the top six ranks. However, the clusters of birds and beetles are
not relevant to the bell shape as predicted by the trained classifier.
Since they are removed from the candidate set in Stage II, the mistake
can be avoided. After the processing of Stage I, the retrieved top 6
samples are all bell shapes.  We will describe the processing in Stages
I and II in detail below.

\subsection{The ICF Stage (Stage I)}

\begin{figure}[t]
\centering
\includegraphics[width=\linewidth]{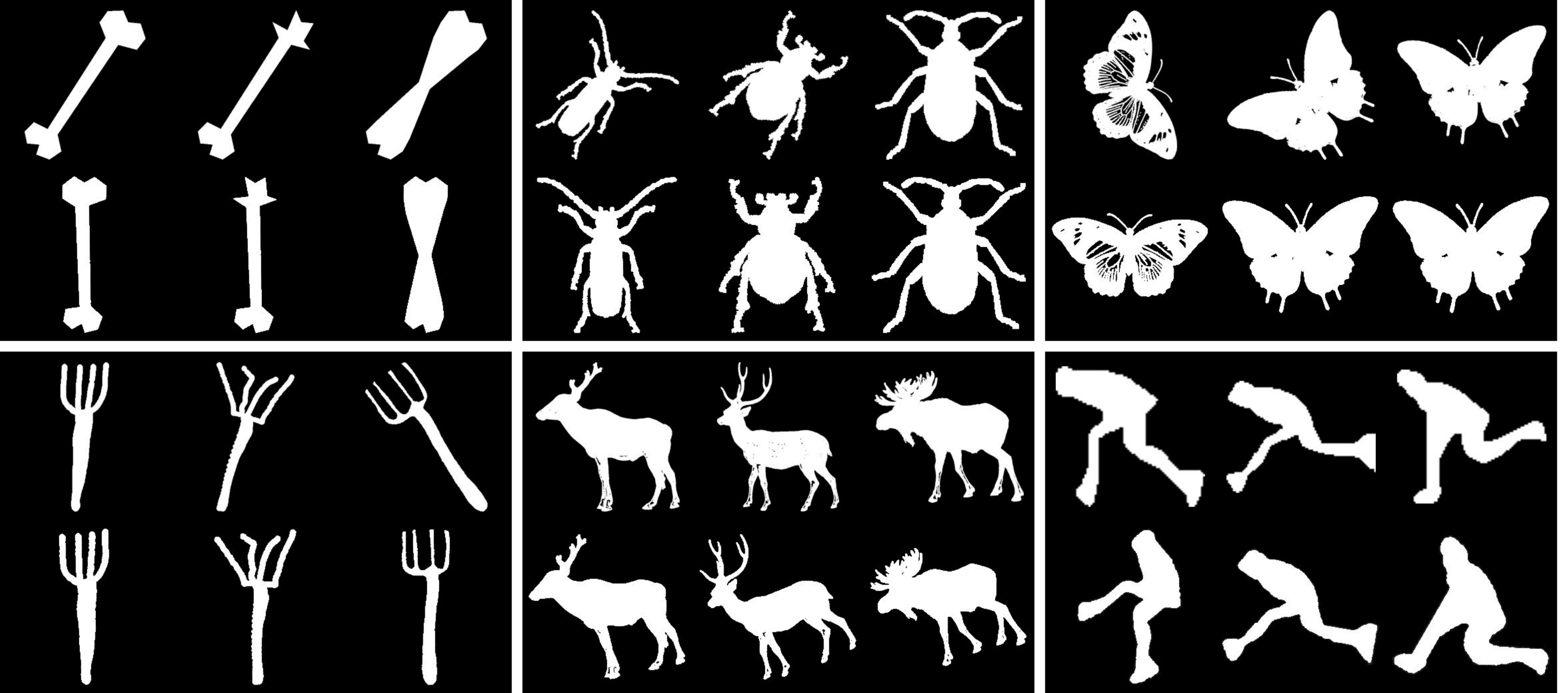}
\caption{Shape normalization results of six classes.} \label{fig.norimg}
\end{figure}

{\bf Shape Normalization.} Each shape is normalized so that it is
invariant against translation, scaling and rotation.  Translational
invariance is achieved by aligning the shape centroid and the image
center.  For rotational invariance, we align the dominant reflection
symmetry axis, which passes through the shape centroid and has the maximum
symmetry value, vertically.  After rotational normalization, we set the
larger side (width or height) of the shape to unity for scale
invariance. Examples of shape normalization are given in Fig.
\ref{fig.norimg}, where the first and third rows are the original shapes
while the second and forth rows are their corresponding normalized
results.  Although normalization based on the dominant reflection
symmetry axis works well in general in our experiments, it is worthwhile
to point out that, if samples of a class do not contain a clear dominant
reflection axis, their normalized poses may not be well aligned (e.g.,
the three running men in the third row of Fig. \ref{fig.norimg}).
However, there still exist rotational invariant global features in TSR
to compensate for the articulational variation.  After shape normalization,
we perform hole-filling and contour smoothing to remove the interior
holes and contour noise, respectively.

{\bf Global Features.} To capture the global layout of a shape, we
consider three feature types: 1) skeleton features, 2) wavelet features,
and 3) geometrical features.

\begin{figure}[t]
\centering
\includegraphics[width=\linewidth]{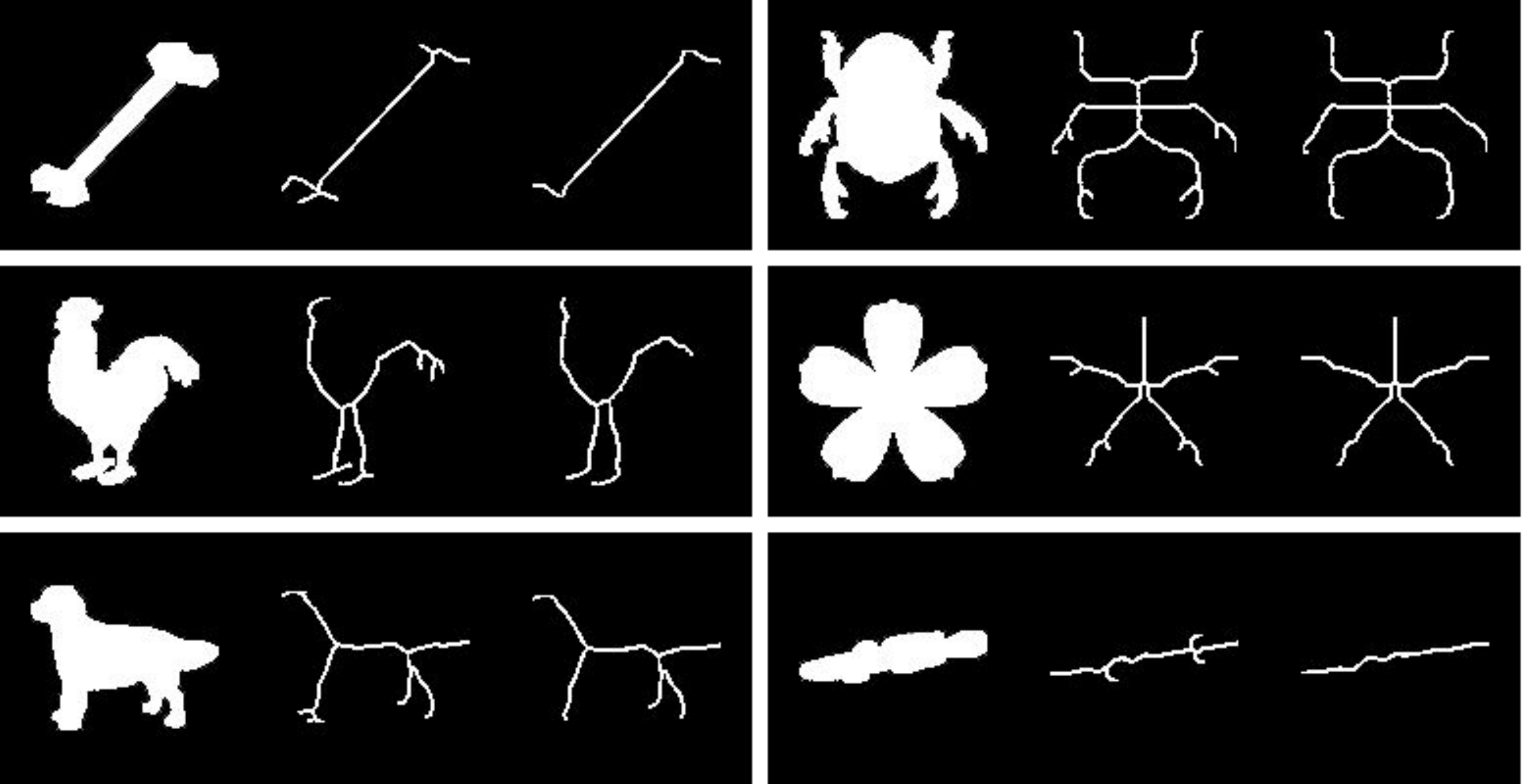}
\caption{Illustration of skeleton feature extraction (from
left to right of each image): the original shape, the initial
skeleton and the pruned skeleton.}\label{fig.skelprune}
\end{figure}

For skeleton features, we extract the basic structural information of a
shape while ignoring minor details due to contour variations. We first
apply the thinning algorithm \cite{Pratt:2007:DIP:1206398} to obtain the initial skeleton. Then, a
pruning process is developed to extract its clean skeleton without
over-simplification. We show several input shapes and their initial and
pruned skeletons in Fig. \ref{fig.skelprune}.  After getting a clean
skeleton, we consider four types of salient points: 1) turning points
(that have a sharp curvature change), 2) end points, 3) T-junction
points and 4) cross-junction points. The numbers of these four salient
points form a 4D skeleton feature vector denoted by $f_s$. Skeleton
features of the six shapes in Fig.  \ref{fig.skelprune} are given in
Table \ref{table.topologicalfeatures}.  They are rotational,
translational and scaling invariant.

\begin{table}[]
\begin{center}
\resizebox{0.9\textwidth}{!}{
\begin{tabular}{|c|c|c|c|c|c|c|}
\hline
Nos. of Salient pts & bone     & beetle   & chicken  & device0  & dog      & fish     \\ \hline
turning pts         & 2        & 4        & 3        & 0        & 1        & 0        \\ \hline
end pts             & 2        & 6        & 4        & 5        & 5        & 2        \\ \hline
T-junction pts      & 0        & 2        & 2        & 0        & 3        & 0        \\ \hline
cross-junction pts  & 0        & 1        & 0        & 1        & 0        & 0        \\ \hline
\end{tabular}}
\end{center}
\caption{Skeleton features for six shapes in Fig.\ref{fig.skelprune}.}
\label{table.topologicalfeatures}
\end{table}

\begin{figure}[t]
\begin{center}
\includegraphics[width=\linewidth]{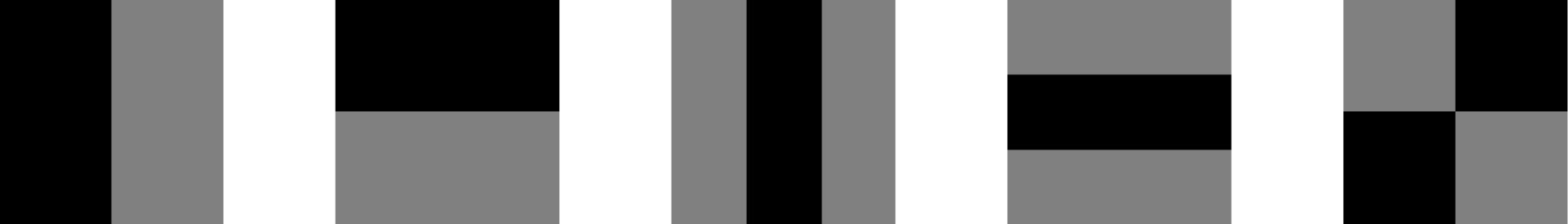}
\caption{Five Haar-like filters used to extract wavelet features
of a normalized shape.}\label{fig.haar}
\end{center}
\end{figure}

For wavelet features, Haar-like filters were adopted for face detection
in \cite{viola2005detecting}. Being motivated by this idea, we adopt
five Haar-like filters as shown in Fig. \ref{fig.haar} to extract
wavelet features. For a normalized shape, the first two filters are used
to capture the 2-band symmetry while the middle two filters are used to
capture the 3-band symmetry horizontally and vertically. The last one is
used to capture the cross diagonal symmetry. The responses of the five
filters form a 5D wavelet feature vector denoted by $f_w$.

Furthermore, we incorporate the following four geometrical features
\cite{Pratt:2007:DIP:1206398}: 1) aspect ratio, 2) circularity, 3)
symmetry and 4) solidity.  The aspect ratio is the ratio of the width
and height of the bounding box of a shape.  The circularity is set to
$4\pi A/P^{2}$, where $A$ is the area and $P$ is the perimeter of the
shape. The aspect ratio and circularity are closer to one, if a shape is
closer to a square or a circle.  If a shape is closer a long bar, its
aspect ratio and circularity are closer to zero. The symmetry is
computed based on the dominant reflection symmetry axis of a shape.  The
solidity of a shape is the ratio of its area and the area of its convex
hull. If a shape is a convex set, its solidity is unity.  Otherwise, it
will be less than one. These four geometric features form a 4D
geometrical feature vector denoted by $f_g$.

\begin{figure}[t]
\begin{center}
\includegraphics[width=\linewidth]{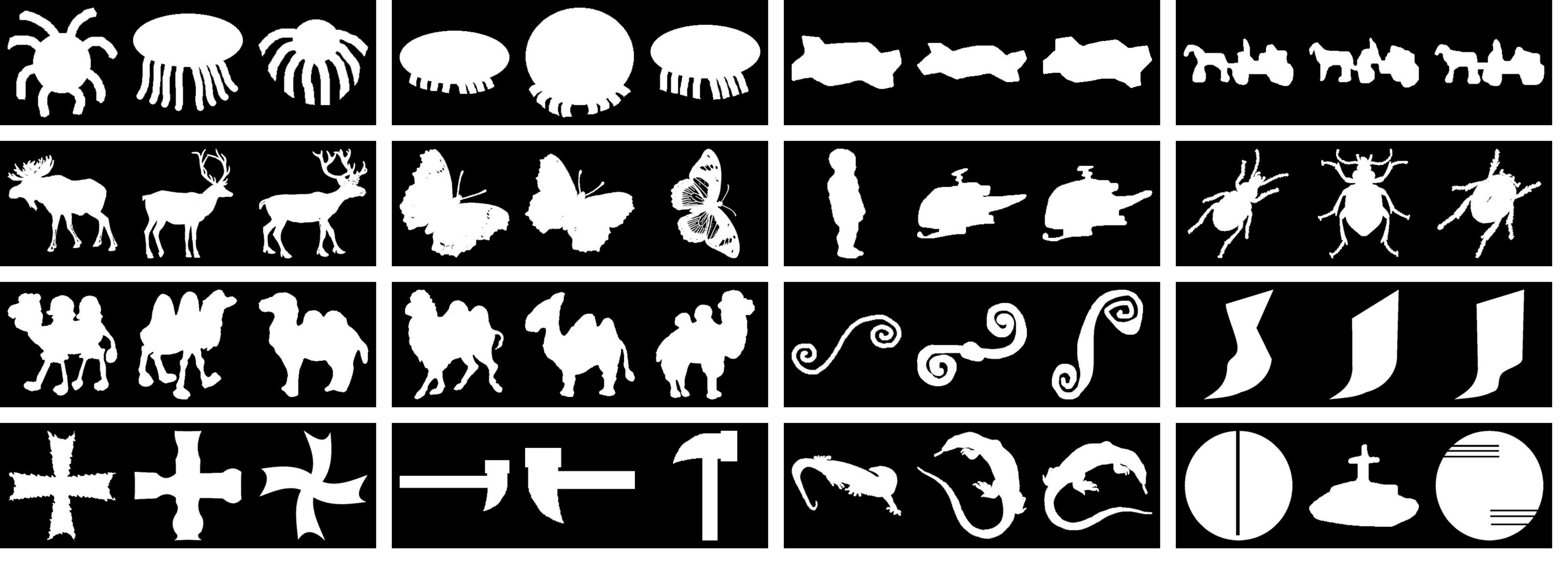}
\caption{Several clustered MPEG-7 dataset shapes using the spectral
clustering method applied in the AIR feature space.}\label{fig.cluster_ex}
\end{center}
\end{figure}

{\bf Shape Clustering.} In the traditional 2D shape retrieval
formulation, all shapes in the dataset are not labeled. Under this
extreme case, we use the spectral clustering algorithm
\cite{ng2002spectral} to reveal the underlying relationship between
gallery shapes based on local features. For the MPEG-7 dataset, shapes
in several clusters using the AIR feature are shown in Fig.
\ref{fig.cluster_ex}. Some clusters look reasonable while others do not.
Actually, any unsupervised clustering method will encounter the
following two challenges.  First, uncertainty occurs near cluster
boundaries so that samples near boundaries have a higher probability of
being wrongly clustered. Second, the total number of shape classes is
unknown.  When the cluster number is larger than the class number in the
database, the clustering algorithm creates sub-classes or even mixed
classes. The relationship between them has to be investigated.

To address the first challenge, we extract $N_i$ samples closest to the
centroid of the $i$th cluster and assign them a cluster label.  Clearly,
samples sharing the same cluster label are close to each other in the
feature space. There is a trade-off in choosing a proper value of $N_i$.
A smaller $N_i$ value guarantees higher clustering accuracy but fewer
gallery samples will be assigned cluster labels.  Empirically, we set
the value of $N_i$ to one half of the size of the $i$th cluster.  To
address the second challenge, we use local features to conduct
clustering and assign cluster labels to a subset of samples. These
labeled samples are used to train a random forest classifier \cite{breiman2001random} with their
global features.  Finally, all gallery shapes are treated as testing
samples. The random forest classifier is used to predict the probability
of each cluster type for them by voting. In this way, samples that are
clustered in the local feature space can be linked to multiple clusters
probabilistically due to similarities in the global feature space.

\begin{figure}[htb]
\begin{center}
\includegraphics[width=\linewidth]{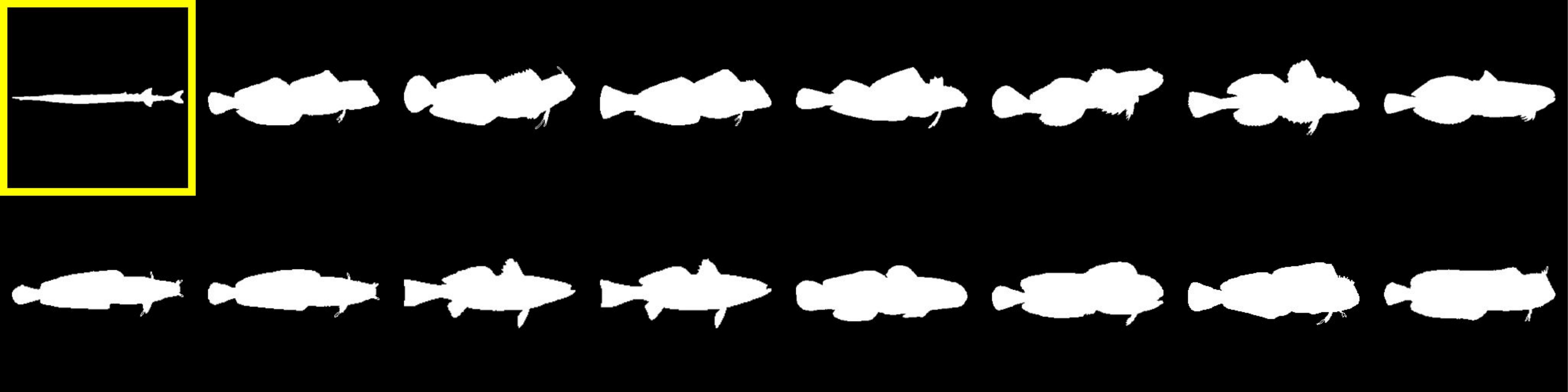}
\caption{Illustration of a shared dominating feature (i.e., the aspect
ratio) among all training fish samples outside of the yellow box, which
could terminate the decision quickly and reject the testing sword-like
fish in the red box.}\label{fig.rfprob}
\end{center}
\end{figure}

Typically, a random forest classifier reduces the influence of outliers
by bootstrapping.  When the training data size is small, it is difficult
to predict outliers in testing samples accurately. It is possible for
the classifier to terminate a decision process in an early stage due to
a shared dominating feature between the testing sample and the training
samples. One example is shown in Fig. \ref{fig.rfprob}, where the
``sword-like" fish sample in the red box is clearly an outlier with
respect to other fish samples. If all fish samples except for the
outlying sample are used as training samples, the aspect ratio is an
important feature for the fish class since its value is consistent among
all training fish samples.  When we use the sword-like fish as a testing
sample, this feature will dominate and terminate the decision process
early with a wrong predicted result. Namely, it would be a sword rather
than a fish.  To overcome this problem, we train multiple random forest
classifiers using different feature subsets to suppress the impact from
a dominating feature. Finally, we combine the results of these
classifiers by the sum rule to obtain the final prediction.

{\bf Cluster Relevance Assignment.} The output of the ICF stage includes:
1) a set of indexed clusters and 2) soft classification (or
multi-labeling) of all gallery samples. For item \#1, we use the
unsupervised spectral clustering algorithm to generate clusters as
described above.  If the class number is known (or can be estimated), it
is desired that the cluster number is larger than the class number.
Each of these clusters is indexed by a cluster ID.  For item \#2, we
adopt soft classification so that each sample can be associated with
multiple clusters. This is done for two reasons. If two sub-classes
belong to the same ground truth class, we need a mechanism to regroup
them together.  Clearly, a hard classification process does not allow
this to happen.  Second, a hard classification error cannot be easily
compensated while a soft classification error is not as fatal and it is
likely to be fixed in the LMR stage (stage II) of the TSR system.

We consider two relevant clusters assignment schemes below.

\noindent
1) \textit{Direct Assignment} \\
We apply the random forest classifier to both training and testing
samples based on their global features. Then, the probability for the
$i$th shape sample (denoted by $y_i$) belongs to the $k$th cluster
denoted by $c_k$ can be estimated by the following normalized voting
result:
\begin{equation} \label{eq.rf_prob}
P_{rf}(y_i \in c_k) = \frac{v_k}{\sum_j{v_j}},
\end{equation}
where $v_k$ is the number of votes claiming that $y_i$ belongs to $c_k$.
Eq. (\ref{eq.rf_prob}) associates $y_i$ to its relevant clusters directly.

\noindent
2) \textit{Indirect Assignment} \\
Intuitively, a good cluster relevance assignment scheme should take both
global and local features into account. This can be achieved as follows.
For query sample $y_i$, we find its $K$ nearest neighbors (denoted by
$x_j$) using a certain distance function in a local feature space (e.g.
the same feature space used in IDSC or AIR).  Then, the probability of
$y_i$ belonging to $c_k$ can be estimated by the weighted sum of the
probability in Eq. (\ref{eq.rf_prob}) in form of
\begin{equation} \label{eq.knn_prob}
P_{knn}(y_i \in c_k) =\! \frac{\sum_{x_j \in knn(y_i)}{P_{rf}(x_j
\in {c_k})}}{\sum_{c_m}{\sum_{x_j \in knn(y_i)}{P_{rf}(x_j \in {c_m})}}}.
\end{equation}
Eq.  (\ref{eq.knn_prob}) associates $y_i$ to its relevant clusters
indirectly. That is, the assignment is obtained by averaging the
relevant clusters assignment of its $K$ nearest neighbors. Empirically,
we choose K to be 1.5 times the average cluster size \ignore{$K=20$} in the experiments.

\begin{figure}[!th]
\centering
\begin{subfigure}[b]{0.8\textwidth}
\centering
\includegraphics[width=0.8\textwidth]{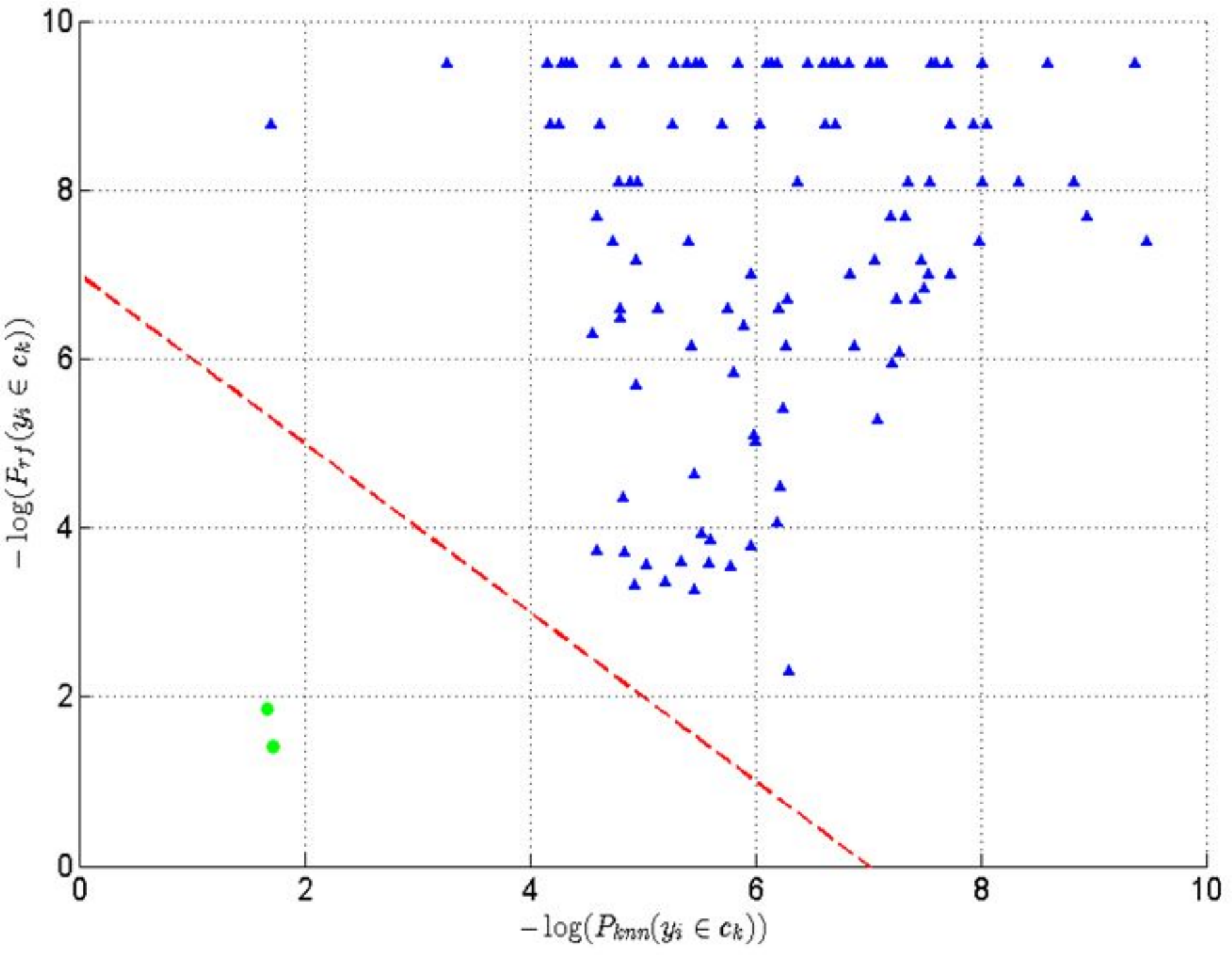}
\caption{}
\end{subfigure}
\centering
\begin{subfigure}[b]{0.8\textwidth}
\centering
\includegraphics[width=0.8\textwidth]{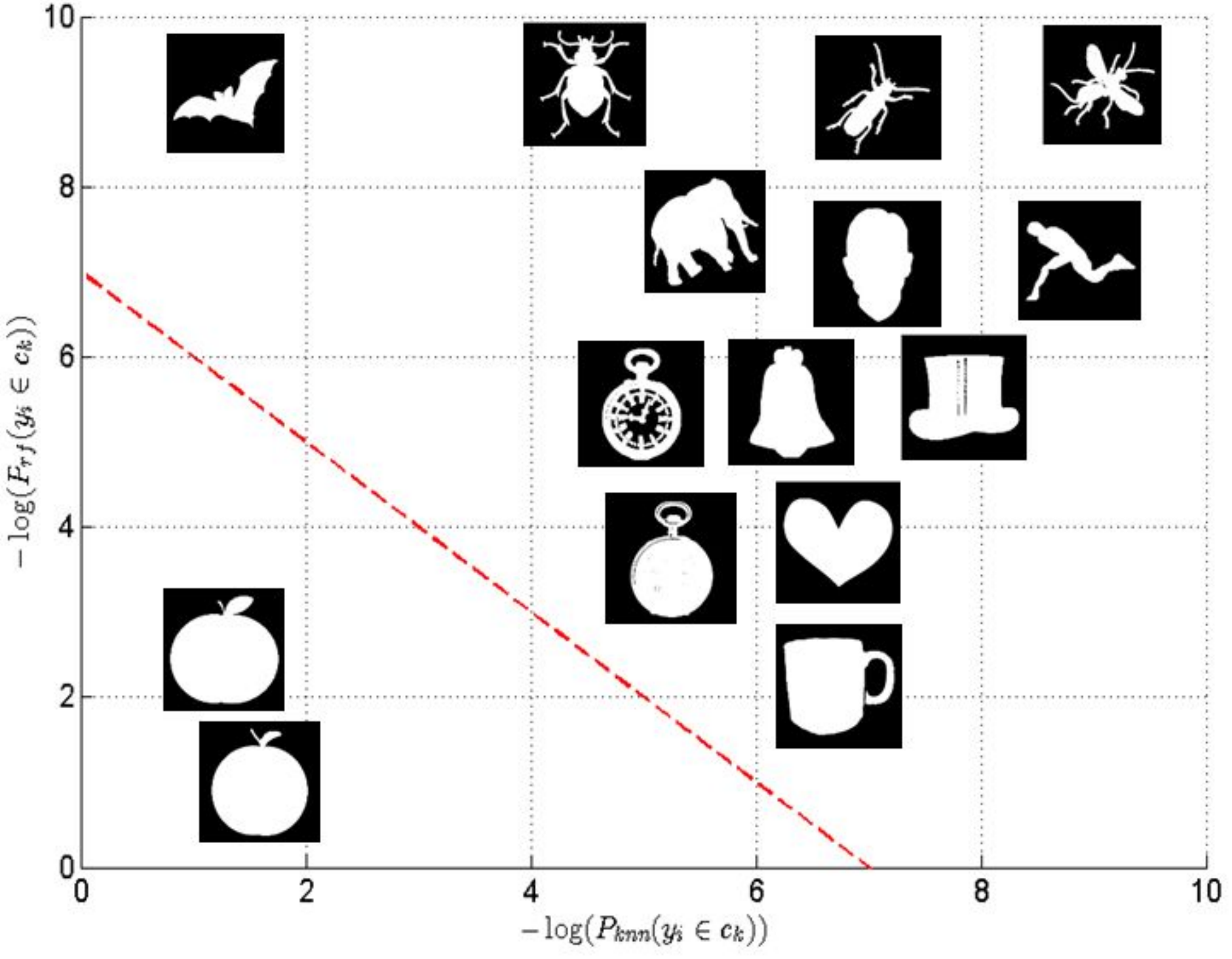}
\caption{}
\end{subfigure}
\caption{Selecting relevant clusters for a query apple shape by
thresholding a cost function as shown in Eq.
(\ref{eq.thresholding}).}\label{fig.costfunction}
\end{figure}

We show an example that assigns a query apple shape to its relevant
clusters in Fig. \ref{fig.costfunction}(a), whose x-axis and y-axis are
the negative log functions of Eqs. (\ref{eq.rf_prob}) and
(\ref{eq.knn_prob}), respectively.  Every dot in Fig.
\ref{fig.costfunction}(a) represents a cluster after the shape
clustering process. To visualize shapes represented by a dot, we \ignore{zoom in
the lower left corner of Fig. \ref{fig.costfunction}(a) and }plot a
representative sample of each cluster in Fig.
\ref{fig.costfunction}(b).

We see that the distance between the bat cluster and the apple cluster
is short in the x-axis but long in the y-axis. This is because that
samples of the apple and bat clusters are interleaved in the local
feature space. This is evidenced in the retrieval results of AIR in Fig.
\ref{fig.irrelevant}(a).  However, the apple and bat clusters have
little intersection in the global feature space. On the other hand, the
cup and apple clusters have large intersection in the global feature
space. Yet, their distance is far in the local feature space. It is
apparent that Eqs. (\ref{eq.rf_prob}) and (\ref{eq.knn_prob}) provide
complementary relevance assignment strategies for query sample $y_i$.
It is best to integrate the two into one assignment scheme. For example,
we can draw a line to separate relevant and irrelevant clusters with
respect to the query apple shape in this plot.

Mathematically, we define a cost function as follows
\begin{align} \label{eq.cost}
\mathbb{J}(y_i,c_k) &= - \log(P_{knn}(y_i \in c_k) P_{rf}(y_i \in c_k)) \nonumber \\
    &= - [\log(P_{knn}(y_i \in c_k)) + \log(P_{rf}(y_i \in c_k))].
\end{align}
We compute $\mathbb{J}(y_i,c_k)$ for all clusters $c_k$. If
\begin{equation}\label{eq.thresholding}
\mathbb{J}(y_i,c_k)<\epsilon,
\end{equation}
where $\epsilon$ is a pre-selected threshold, we say that cluster $c_k$ is a
relevant cluster for query $y_i$. Otherwise, it is irrelevant.

\subsection{The LMR Stage (Stage II)}

In the LMS stage, we rank the similarity of shapes in the retrieved
relevant clusters by using a local-features-based matching scheme (e.g.,
AIR) including a diffusion process. We adopt the Local Constrained
Diffusion Process (LCDP) from \cite{donoser2013diffusion} in the TSR
system. The diffusion process is slightly modified with the availability
of relevant clusters in the TSR system since the diffusion process can
be conducted on a more reasonable manifold due to the processing in
Stage I. 

\section{Experimental Results} \label{sec.experiment}

We demonstrate the retrieval performance of the proposed TSR system by
conducting experiments on three shape datasets: MPEG-7, Kimia99 and
Tari1000. We set the threshold $\epsilon$ of the cost function in Eq.
(\ref{eq.thresholding}) to 7 empirically for all experiments. We also
consider the incorporation of two diffusion processes in various
schemes. They are denoted by:
\begin{itemize}
\itemsep -1ex
\item DP1: the diffusion process proposed in \cite{yang2009locally},
\item DP2: the diffusion process proposed in \cite{donoser2013diffusion}.
\end{itemize}

\noindent{\textbf {MPEG-7 Shape Dataset}.} The MPEG-7 shape dataset
\cite{latecki2000shape}, which remains to be the largest and most
challenging, contains 1400 shape samples in 70 independent classes.
Samples are uniformly distributed so that each class has $C=20$ shape
samples. The retrieval performance is measured by the bull's eye score,
which means the percentage of shapes sharing the same class with a query
in the top $2 \times C=40$ retrieved shapes.  AIR and DP1 are used as
the local feature space and the diffusion process in the TSR method,
respectively. 

\begin{table*}[!ht]
\begin{center}
\resizebox{\linewidth}{!}{
\begin{tabular}{|c|c|c|c|c|c|c|c|c|}
\hline
Cluster Numbers (M) & 16      & 32      & 48
                       & 64      & 80      & 96      & 112     & 128   \\ \hline
TSR (ICF+AIR)       & 96.00\% & 97.54\% & 98.81\%
                       & 99.51\% & 99.62\% & 99.85\% & \textbf{99.92\%} & 99.90\% \\ \hline
TSR (ICF+AIR+DP1)       & 98.32\% & 98.90\% & 99.72\%
                       & 99.99\% & 99.99\% & 99.99\% & \textbf{100.00\%} & 99.99\% \\ \hline
\end{tabular}}
\end{center}
\caption{Comparison of bull's eye scores with different cluster numbers
for the TSR method.}\label{table.score_on_M}
\end{table*}
We first show the bull's eye scores of two TSR schemes using a different
cluster number $M$ in the shape clustering step in Table
\ref{table.score_on_M}. Both TSR methods adopt the AIR features for the
distance computation. However, one of them uses DP1 while the other does
not. Since TSR(ICF+AIR+DP1) offers better performance, we choose it as
the default TSR configuration for the MPEG-7 shape dataset. Generally
speaking, the performance degrades when $M$ is small due to the loss of
discriminability in larger cluster sizes. The retrieval performance
improves as the cluster number increases up to 112. After that, the
performance saturates and could even drop slightly. That means that we
lose the advantage of clustering when the cluster size is too small.
For the remaining MPEG-7 dataset experimental results, we choose
$M=112$.

The bull's eye scores of the TSR method and several state-of-the-art
methods are compared in Table \ref{table.performance_2}. Both TSR and
AIR+DP2 reach 100\%.  The bull's eye score is one of the popular
retrieval performance measures for the 2D shape retrieval problem.

\begin{table}[th]
\begin{center}
\resizebox{0.8\textwidth}{!}{
\begin{tabular}{|l|c|}
\hline
Method                  & Bull's eye score \\ \hline
CSS \cite{mokhtarian1997efficient}                     & 75.44\%          \\ \hline
IDSC \cite{ling2007shape}                   & 85.40\%          \\ \hline
ASC  \cite{ling2010balancing}                   & 88.30\%          \\ \hline
HF  \cite{wang2012shape}                    & 89.66\%          \\ \hline
AIR  \cite{gopalan2010articulation}                   & 93.67\%          \\ \hline
IDSC+DP1 \cite{yang2009locally}              & 93.32\%          \\ \hline
ASC+DP1  \cite{ling2010balancing}              & 95.96\%          \\ \hline
IDSC+SC+Co-Transduction \cite{bai2012co} & 97.72\%          \\ \hline
AIR+TPG \cite{yang2013affinity}                & 99.90\%           \\ \hline
AIR+DP2 \cite{donoser2013diffusion}                  & 100.00\%            \\ \hline
Proposed TSR (ICF+AIR+DP1)    & \bf{100.00\%}            \\ \hline
\end{tabular}}
\end{center}
\caption{Comparison of bull's eye scores of several state-of-the-art methods
for the MPEG-7 dataset.}\label{table.performance_2}
\end{table}
\begin{table}[th]
\begin{center}
\resizebox{\linewidth}{!}{
\begin{tabular}{|c|c|c|c|c|c|}
\hline
N           & 20               & 25               & 30               & 35               & 40             \\ \hline
IDSC         & 77.21\%          & 80.44\%          & 82.61\%          & 84.16\%          & 85.40\%        \\ \hline
IDSC+DP1     & 88.53\%          & 90.78\%          & 92.03\%          & 92.73\%          & 93.32\%        \\ \hline
AIR          & 88.17\%          & 89.99\%          & 91.28\%          & 92.64\%          & 93.67\%        \\ \hline
AIR+DP2      & 94.42\%          & 97.92\%          & 98.66\%          & 99.38\%          & 100\%          \\ \hline
Proposed TSR  & \textbf{98.46\%} & \textbf{99.09\%} & \textbf{99.40\%} & \textbf{99.71\%} & \textbf{100\%} \\ \hline
\end{tabular}}
\end{center}
\caption{Comparison of top 20, 25, 30, 35, 40 retrieval accuray for MPEG-7 dataset.}
\label{table.topknum}
\end{table}

However, since each MPEG-7 shape class contains 20 shape samples, the
measure of correctly retrieved samples from the top 40 ranks cannot
reflect the true power of the proposed TSR method. To push the retrieval
performance further, we compare the accuracy of retrieved results from
the top 20, 25, 30, 35 and 40 ranks of TSR and several state-of-the-art
methods in Table \ref{table.topknum} whose last column corresponds the
bull's eye scores reported in Table \ref{table.performance_2}.  The
superiority of TSR stands out clearly in this table.
\begin{figure}[!h]
\centering
\begin{subfigure}[b] {0.45\linewidth}
\centering
\includegraphics[width=\linewidth]{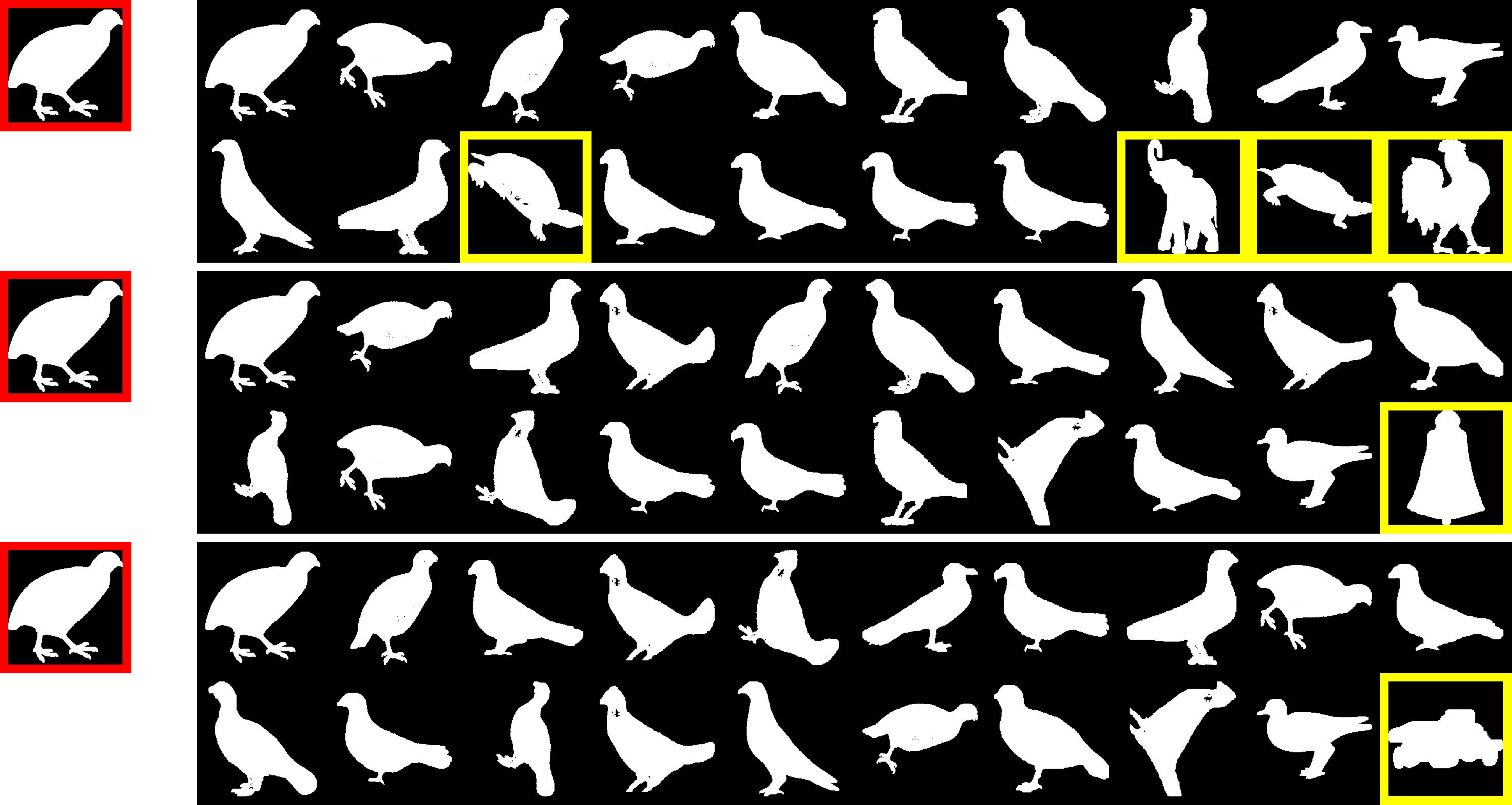}
\caption{Bird}
\end{subfigure}
\begin{subfigure}[b]{0.45\linewidth}
\centering
\includegraphics[width=\linewidth]{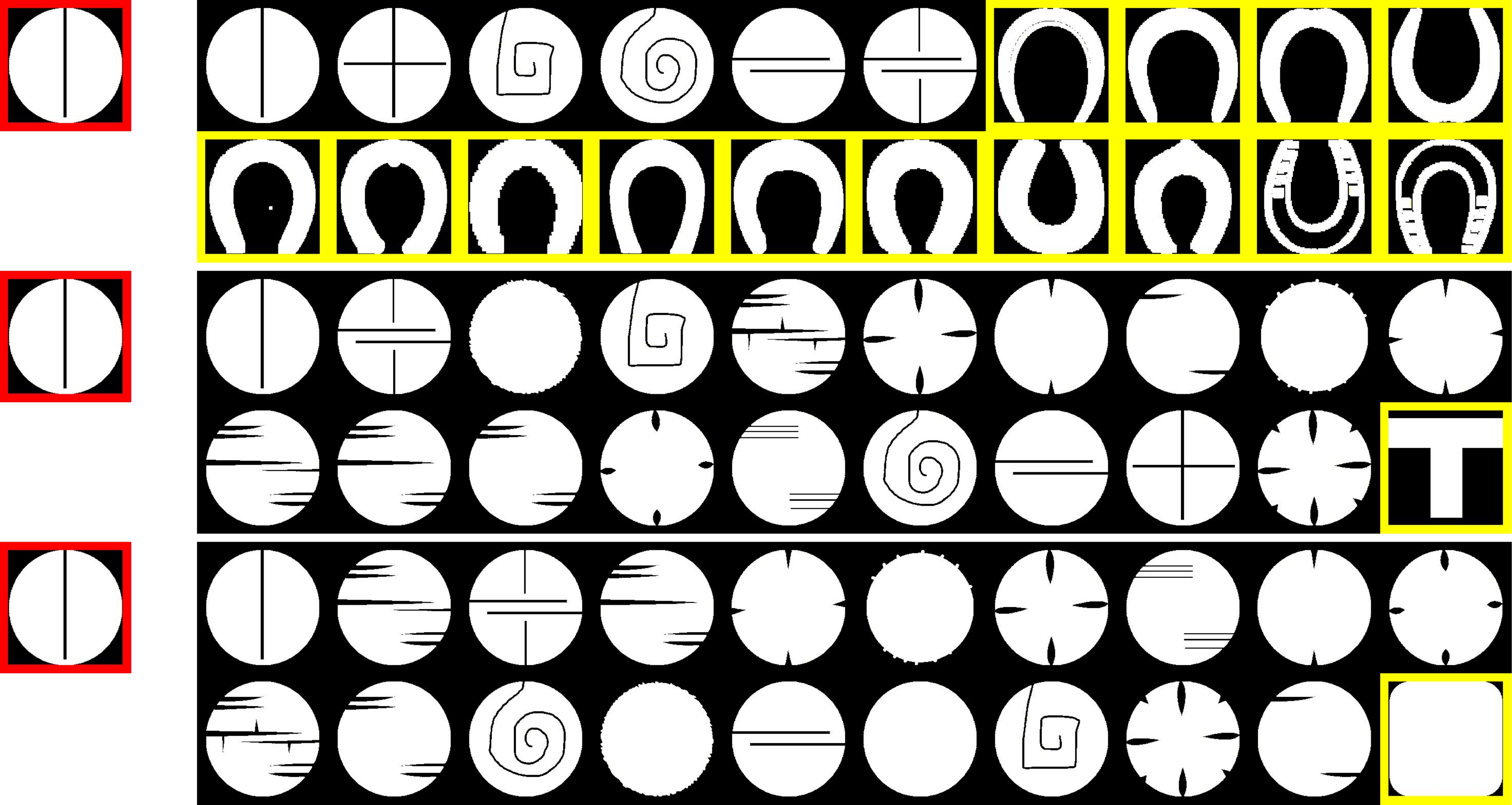}
\caption{Device9}
\end{subfigure}
\begin{subfigure}[b]{0.45\linewidth}
\centering
\includegraphics[width=\linewidth]{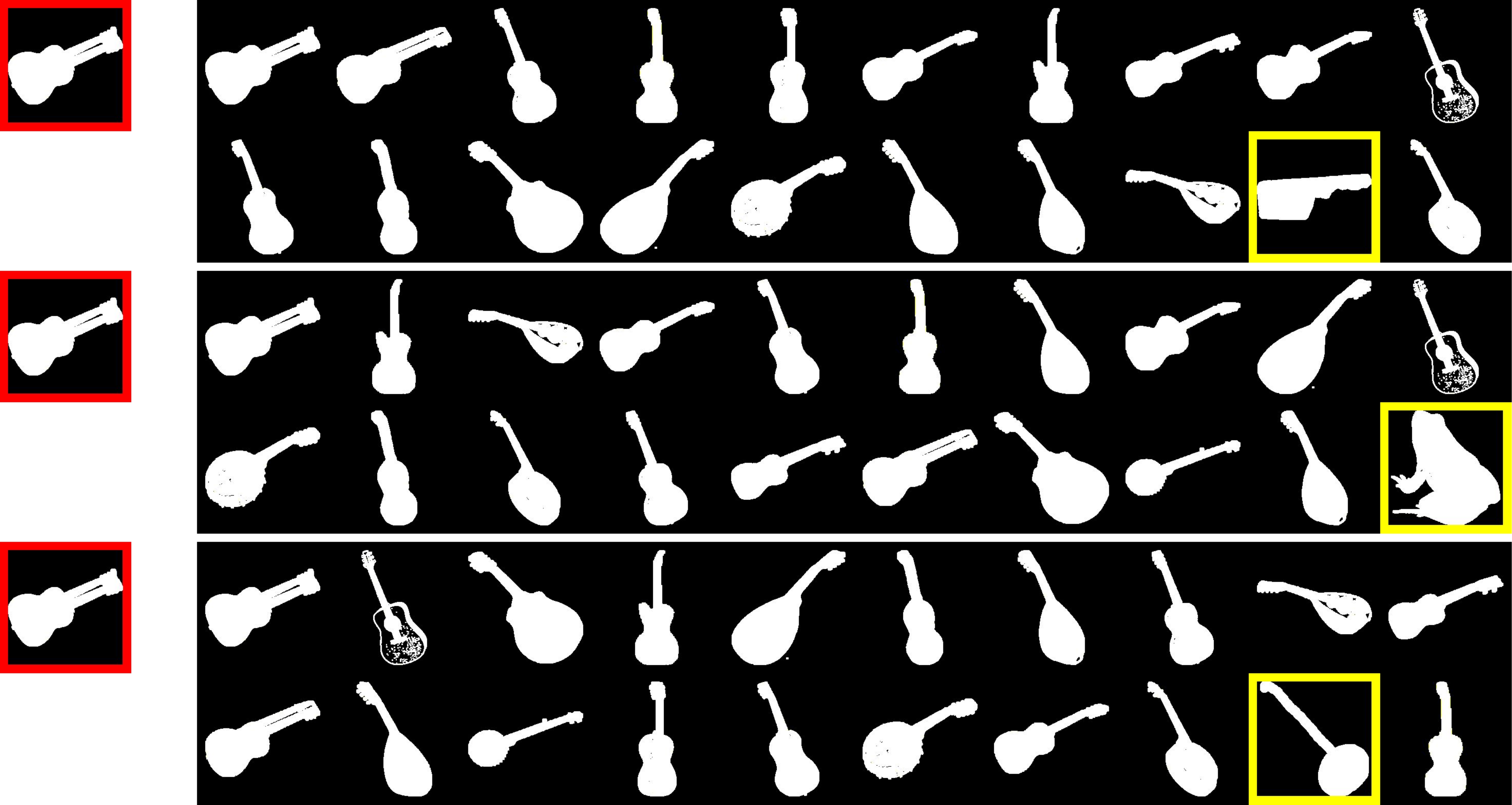}
\caption{Guitar}
\end{subfigure}
\begin{subfigure}[b]{0.45\linewidth}
\centering
\includegraphics[width=\linewidth]{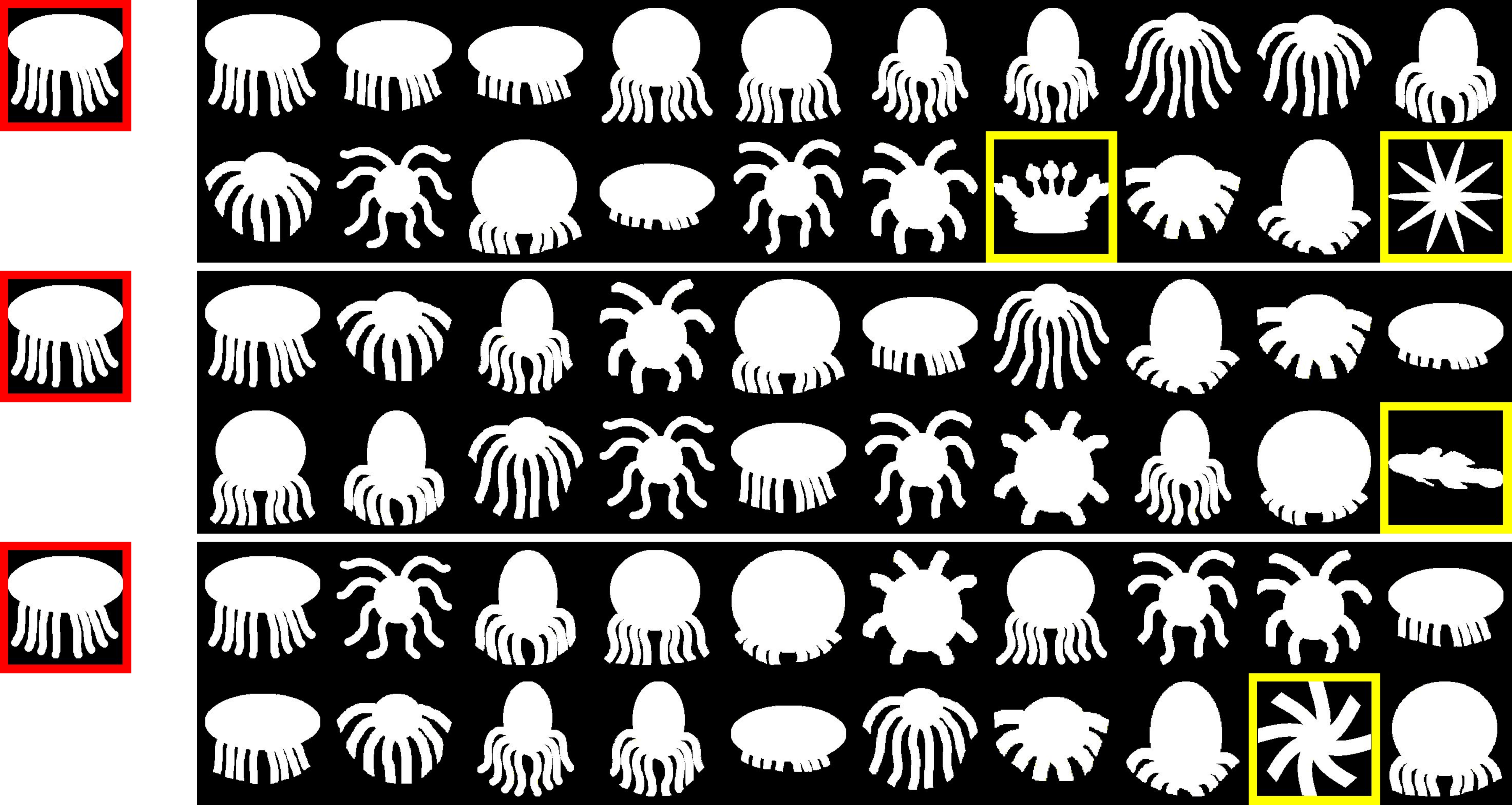}
\caption{Octopus}
\end{subfigure}
\caption{Comparison of retrieved rank-ordered shapes (left-to-right in
the top row followed by left-to-right in the second row within each
black stripe). For each query case, retrieved results of IDSC+DP1,
AIR+DP2 and TSR are shown in the first, second and third black stripes
of all subfigures, respectively.} \label{fig.errors}
\end{figure}

When $N=20$, TSR can retrieve 20 shapes of the entire class correctly
with respect to most query samples. However, it still makes mistakes
occasionally. It is worthwhile to show these erroneous cases to have
further insights.  For this reason, we conduct error analysis in Figs.
\ref{fig.errors}(a)-(d). The performance of IDSC+DP1 is clearly worse
than that of AIR+DP2 and TSR. AIR+DP2 makes mistakes between bird/bell,
circle/device8, guitar/frog and octopus/fish as shown in the
second black stripes of all subfigures. This type of mistakes is not
consistent with human visual experience. In contrast, TSR makes mistakes
between bird/truck , device9/device3, guitar/spoon, octopus/device2 as shown in
the third black stripes of all subfigures. These mistakes are closer to
human visual experience. Actually, these wrongly retrieved shapes are
similar to the query shape in their global attributes as a result of the
special design of the TSR system.
\begin{figure}[!h]
\centering
\includegraphics[width = 0.9\linewidth]{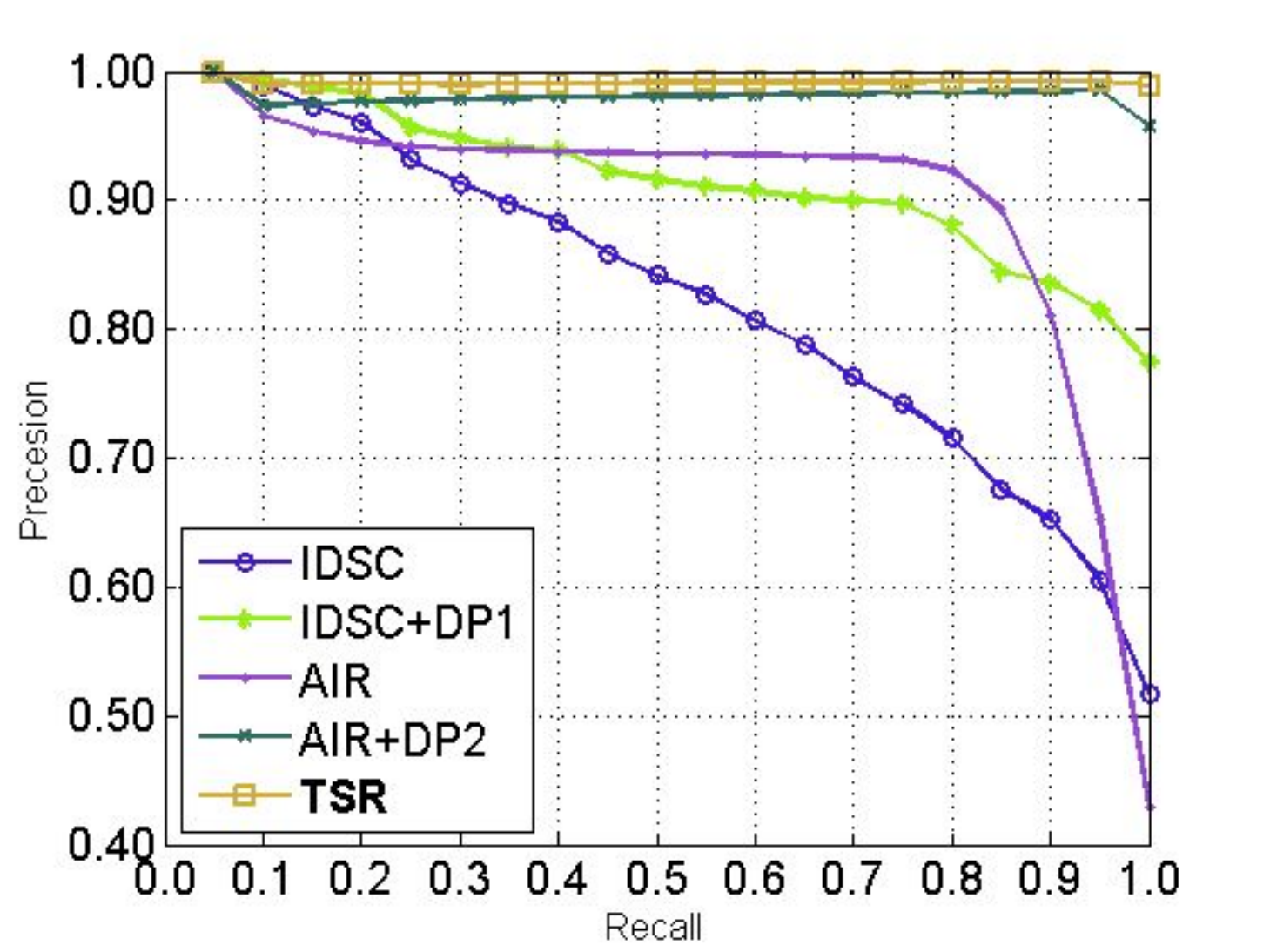}
\caption{Comparison of precision-and-recall curves of several methods
for the MPEG-7 dataset.}\label{fig.topn}
\end{figure}
For further performance benchmarking, we show the precision-and-recall
curves of TSR and several methods in Fig.  \ref{fig.topn}.  We see from
the figure that TSR outperforms all other methods by a significant
margin except for AIR+DP2.

\noindent{\textbf{Kimia99 Shape Dataset}.} The MPEG-7 shape dataset
contains primarily articulation and contour deformation variations. We
also conduct experiments on the Kimia99 shape dataset
\cite{sebastian2004recognition} that contains other variations such as
occlusions and distorted parts. However, this dataset is relatively
small. It contains 99 shapes in total, which are classified into 9
classes. Each class has 11 shapes.  For this dataset, we choose IDSC as
the local feature extraction and ranking method and DP1 as the diffusion
process in the LMR stage as the default TSR method.  The number of
clusters is set to 15.

\begin{table}[!h]
\begin{center}
\resizebox{\textwidth}{!}{
\begin{tabular}{|l|l|l|l|l|l|l|l|l|l|l|}
\hline
Method                                  & 1st      & 2nd      & 3rd      & 4th      & 5th      & 6th      & 7th      & 8th      & 9th      & 10th     \\ \hline
SC\cite{belongie2002shape}              & 97       & 91       & 88       & 85       & 84       & 77       & 75       & 66       & 56       & 37       \\ \hline
Gen. Model \cite{tu2004shape}           & 99       & 97       & 99       & 98       & 96       & 96       & 94       & 83       & 75       & 48       \\ \hline
Path Similarity \cite{bai2008path}      & 99       & 99       & 99       & 99       & 96       & 97       & 95       & 93       & 89       & 73       \\ \hline
Shock Edit \cite{sebastian2004recognition} & 99       & 99       & 99       & 98       & 98       & 97       & 96       & 95       & 93       & 82       \\ \hline
Triangle Area \cite{alajlan2008geometry}   & 99       & 99       & 99       & 98       & 98       & 97       & 97       & 98       & 94       & 79       \\ \hline
Shape Tree \cite{felzenszwalb2007hierarchical}                             & 99       & 99       & 99       & 99       & 99       & 99       & 99       & 97       & 93       & 86       \\ \hline
IDSC\cite{ling2007shape}                & 99       & 99       & 99       & 98       & 98       & 97       & 97       & 98       & 94       & 79       \\ \hline
IDSC+GT\cite{bai2010learning}           & 99       & 99       & 99       & 99       & 99       & 99       & 99       & 99       & 97       & 99       \\ \hline
Proposed TSR & {\bf 99} & {\bf 99} & {\bf 99} & {\bf 99} & {\bf 99} & {\bf 99} & {\bf 99} & {\bf 99} & {\bf 99} & {\bf 99} \\ \hline
\end{tabular}}
\end{center}
\caption{Comparison of top N consistency of several shape retrieval
methods for the Kimia99 dataset.}\label{table.performance_kimia}
\end{table}

\begin{figure}[!h]
\centering
\includegraphics[width = 0.9\linewidth]{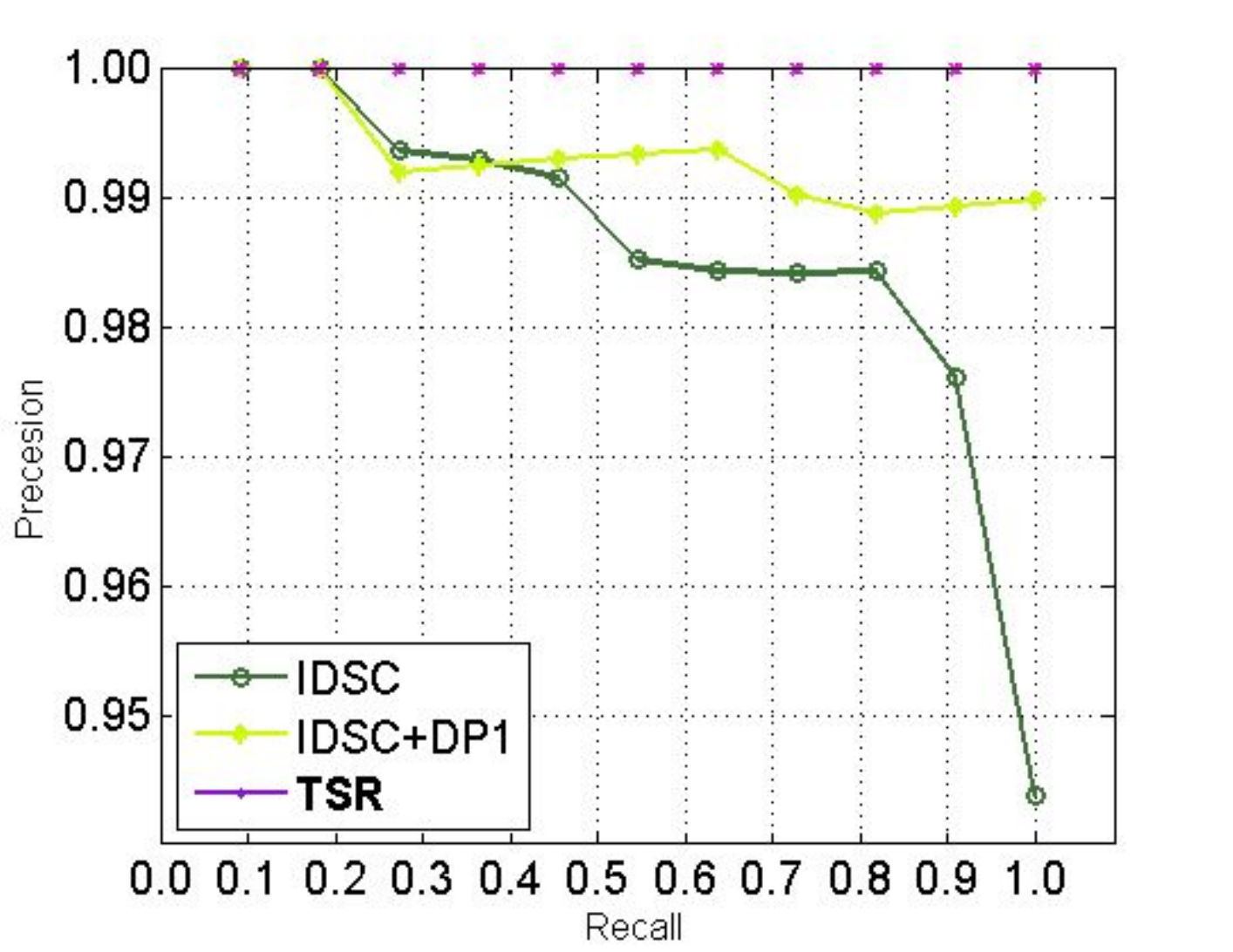}
\caption{Comparison of the precision and recall curves of several
shape retrieval methods for the Kimia99 dataset.}\label{fig.topn_kimia}
\end{figure}

The common evaluation criterion for this dataset is top $N$ (with $N=
1,2, \cdots, 10$) consistency, which measures consistency at the top
$N^{th}$ retrieved shapes against each query. Note that the best
possible value is 99, which is summed up by consistency of all 99 query
samples. The top $N$ consistency results of several methods are compared
in Table \ref{table.performance_kimia}.  The TSR method can exclude
round 75\% irrelevant shapes for each query and effectively improve the
IDSC result. The TSR method can achieve the highest consistency, namely,
99, for all possible $N$ values. The precision-and-recall curves of
IDSC, IDSC+DP1 and TSR are shown in Fig. \ref{fig.topn_kimia}. We
conclude that the TSR method does not make any mistake in shape
retrieval against the Kimia99 Shape Dataset as supported by data in
Table \ref{table.performance_kimia} and Fig. \ref{fig.topn_kimia}.
\begin{table}[htb]
\begin{center}
\resizebox{0.8\linewidth}{!}{
\begin{tabular}{|l|c|}
\hline
Method                        & Bull's eye score \\ \hline
SC \cite{belongie2002shape}                           & 94.17\%          \\ \hline
IDSC \cite{ling2007shape}                         & 95.33\%          \\ \hline
ASC \cite{ling2010balancing}                           & 95.44\%          \\ \hline
IDSC+GT \cite{bai2010learning}                      & 99.35\%          \\ \hline
IDSC+LCDP \cite{bai2012co}                    & 99.70\%          \\ \hline
IDSC+DDGM+Co-Transduction \cite{bai2012co}    & 99.995\%         \\ \hline
Proposed TSR & \bf{100.00\%}            \\ \hline
\end{tabular}}
\end{center}
\caption{Comparison of bull's eye scores of several shape retrieval methods
for the Tari1000 dataset.}\label{table.performance_Tari}
\end{table}

\begin{figure}[!h]
\centering
\includegraphics[width = 0.9\linewidth]{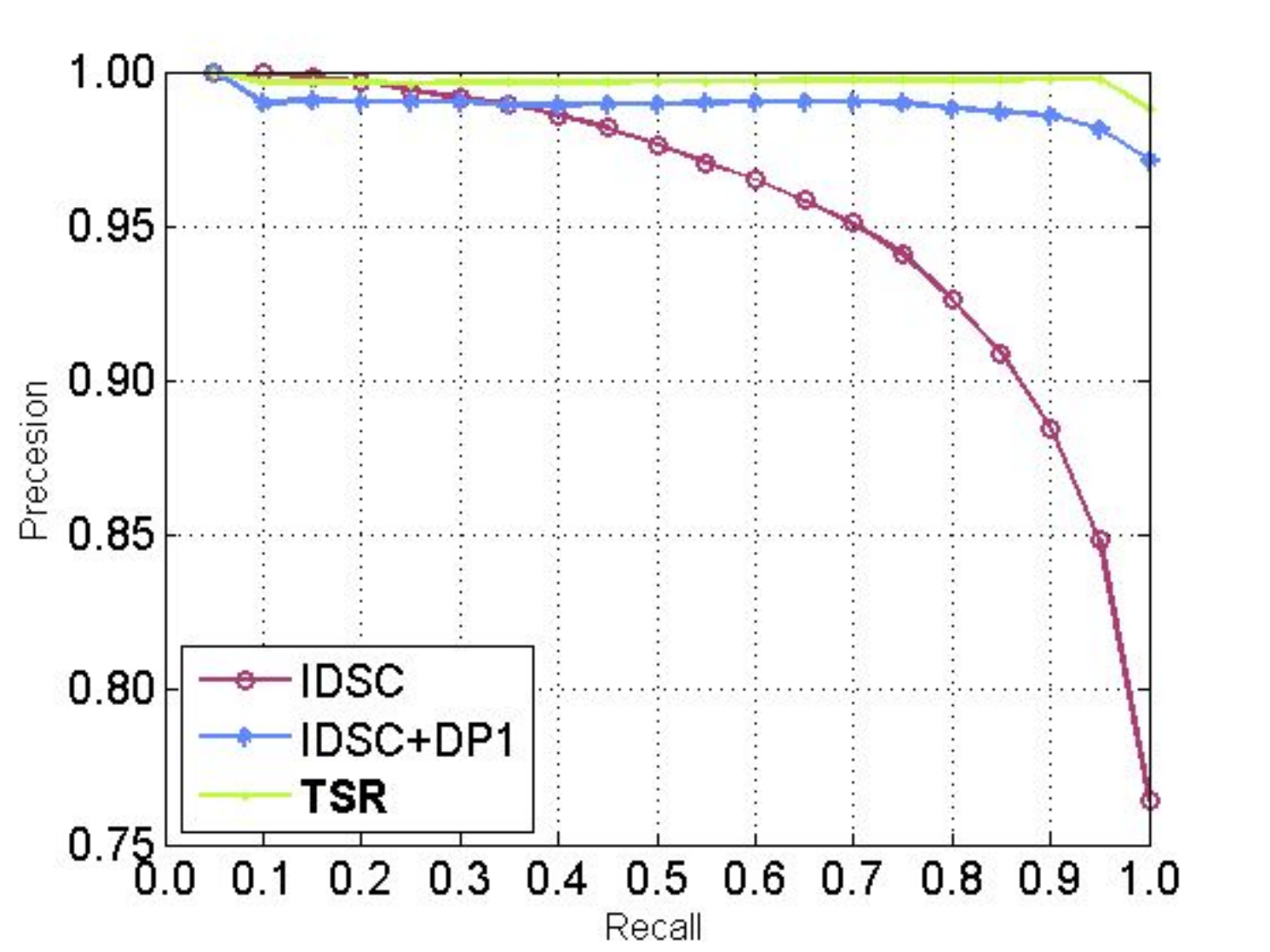}
\caption{Comparison of precision and recall curves of several shape retrieval
methods for the Tari1000 dataset.}\label{fig.topn_tari}
\end{figure}

\noindent{\textbf {Tari1000 Shape Dataset}.} We test the TSR method on a
new dataset called Tari1000 \cite{aslan2008disconnected}. Tari1000
consists 1000 shapes classified into 50 classes. Each class has 20
shapes. As compared to the MPEG-7 dataset, Tari1000 contains more
deformation and articulation variations. Here, we adopt IDSC as the
local feature extraction and ranking method and DP1 as the diffusion
process in the LMR stage as the default TSR method.  The number of
clusters is set to 75. The bull's eye scores of several methods are compared in Table
\ref{table.performance_Tari} and the Precision and Recall curves are
shown in Fig. \ref{fig.topn_tari}. We see that TSR can achieve the
perfect Bull's eye score (100\%) which proves its robustness against
severe articulations.

\section{Conclusion} \label{sec.conclusion}

A robust two-stage shape retrieval (TSR) method was proposed to solve
the 2D shape retrieval problem. In the ICF stage, the TSR method
explores the underlying global properties of 2D shapes. Irrelevant shape
clusters are removed for each query shape. In the LMR stage, the TSR
method only need to focus on the matching and ranking in a much smaller
subset of shapes. We conducted thorough retrieval performance evaluation
on three popular datasets - MPEG7, Kimia99 and Tari1000.  The TSR method
retrieves more globally similar shapes and achieves the highest
retrieval accuracy as compared with its benchmarking methods.

\section*{Acknowledgment}

Computation for the work described in this paper was supported by the
University of Southern California's Center for High-Performance
Computing (hpc.usc.edu).



\bibliographystyle{elsarticle-num} 
\bibliography{ArXiv_2D}





\end{document}